# IDENTIFYING UNFAIR ITEMS IN EDUCATIONAL MEASUREMENT

Yefim Bakman

Tel-Aviv University, *E-mail:* bakman@post.tau.ac.il

**Abstract**. Measurement professionals cannot come to an agreement on the definition of the term 'item fairness'. In this paper a continuous measure of item unfairness is proposed. The more the unfairness measure deviates from zero, the less fair the item is. If the measure exceeds the cutoff value, the item is identified as definitely unfair. The new approach can identify unfair items that would not be identified with conventional procedures. The results are in accord with experts' judgments on the item qualities. Since no assumptions about scores distributions and/or correlations are assumed, the method is applicable to any educational test. Its performance is illustrated through application to scores of a real test.

All researchers without exceptions confess the existence of unfair items, nevertheless Cole & Zieky (2001) indicated: 'there is no generally accepted definition of fairness with respect to testing'. Indeed, the definition should comprise a method for the differentiation between fair and unfair items, a method which is based on some measure. If an appropriate method existed, it would provide a definition of unfair items.

Some measurement professionals use the term 'unfair item' in narrow sense implying that such item favors one group of test-takers over another even though the groups are of comparable ability. Differential item functioning method (DIF) verifies whether a test item is biased against particular gender, ethnic, social, or economic groups (e.g. Holland, 1988). DIF tries to assess item functioning by comparison of the item scores for two or more groups. DIF cannot check item fairness on a single group; consequently it cannot present a fairness scale.

Speaking about unfair items in broad sense, including ambiguous items and faulty teaching, then item analysis is employed which identifies unfair items with the help of the item discrimination index and item difficulty. Here again an exact measure of item fairness is absent. For example, item analysis cannot tell which of two items is fairer: whose discrimination index equals 0.3 or 0.5? It is also impossible to say which items are ideally fair. Researchers (Zimmerman & Zumbo, 1993; Zimmerman & Williams, 1980; Burton, 2001, 2006; Harrison, 1986) subjected to criticism the conventional methods of test analysis.

In what follows, we propose a measure of item unfairness. The method represents each item by means of two parameters, which may be interpreted as coordinates of a point in a plane. In such geometrical presentation fair items are localized near the so-called "ideal line", whereas unfair items are lying far from it. The farther an item point is from the ideal line, the less fair the corresponding item is. Thus, the distance from the ideal line presents the unfairness measure sought for. The new approach can identify unfair items that would not be identified with existing procedures.

## *1. Method of item representation*

Firstly, we will consider an educational test whose items are dichotomously scored: 1- right, 0 - wrong. It will be shown subsequently that this restriction may be easily taken off and the method applies to tests in which some responses receive fractional credit as well. Table 1 shows a test outcome, where $x_{ki} = 1$ or 0 for dichotomous scores.

| Person | Item | | | | | |
|---|---|---|---|---|---|---|
| | 1 | 2 | … | i | … | N |
| 1 | $x_{11}$ | $x_{12}$ | … | $x_{1i}$ | … | $x_{1N}$ |
| 2 | $x_{21}$ | $x_{22}$ | … | $x_{2i}$ | … | $x_{2N}$ |
| … | … | … | … | … | … | … |
| k | $x_{k1}$ | $x_{k2}$ | … | $x_{ki}$ | … | $x_{kN}$ |
| … | … | … | … | … | … | … |
| K | $x_{K1}$ | $x_{K2}$ | … | $x_{Ki}$ | … | $x_{KN}$ |

Table 1. Person by Item matrix containing scores for each examinee on each item.

The method of test analysis is as follows: an equation of linear regression expressing the proportion of correct responses on personal scores is calculated for each item. To illustrate the method we will consider a real example in which a total of 250 students took a multiple-choice test consisting of 40 items. All items were equally weighted (2.5 points for each correct choice) giving the maximum grade of 100 points. Since the instructional pattern might influence the scores, we will begin with discussion of a group of 55 students whose lecturer was B.

Four typical items were chosen by means of experts' judgments, one for each type: easy, moderate, hard, and unfair item. The classification was also based on the number of correct responses. With such a criterion unfair items are similar to the difficult ones (few correct responses), however during the examination the students were allowed to ask questions about unclear or ambiguous items, those appeals helped to distinguish between hard and unfair items.

The presentation of the items by means of the linear regression lines is shown in Fig.1, where the four exemplars, one for each of the four item types, are presented. From these instances it is seen that the regression line is almost horizontal for easy Item 32, it is steep for hard Item 24 – most of students which scored poorly replied incorrectly. It is interesting that the regression line of unfair Item 5 is similar to that of easy Item 32 as to the slope, but passes low since many students responded incorrectly independently of their overall knowledge. The analysis does not make allowance for guessing.

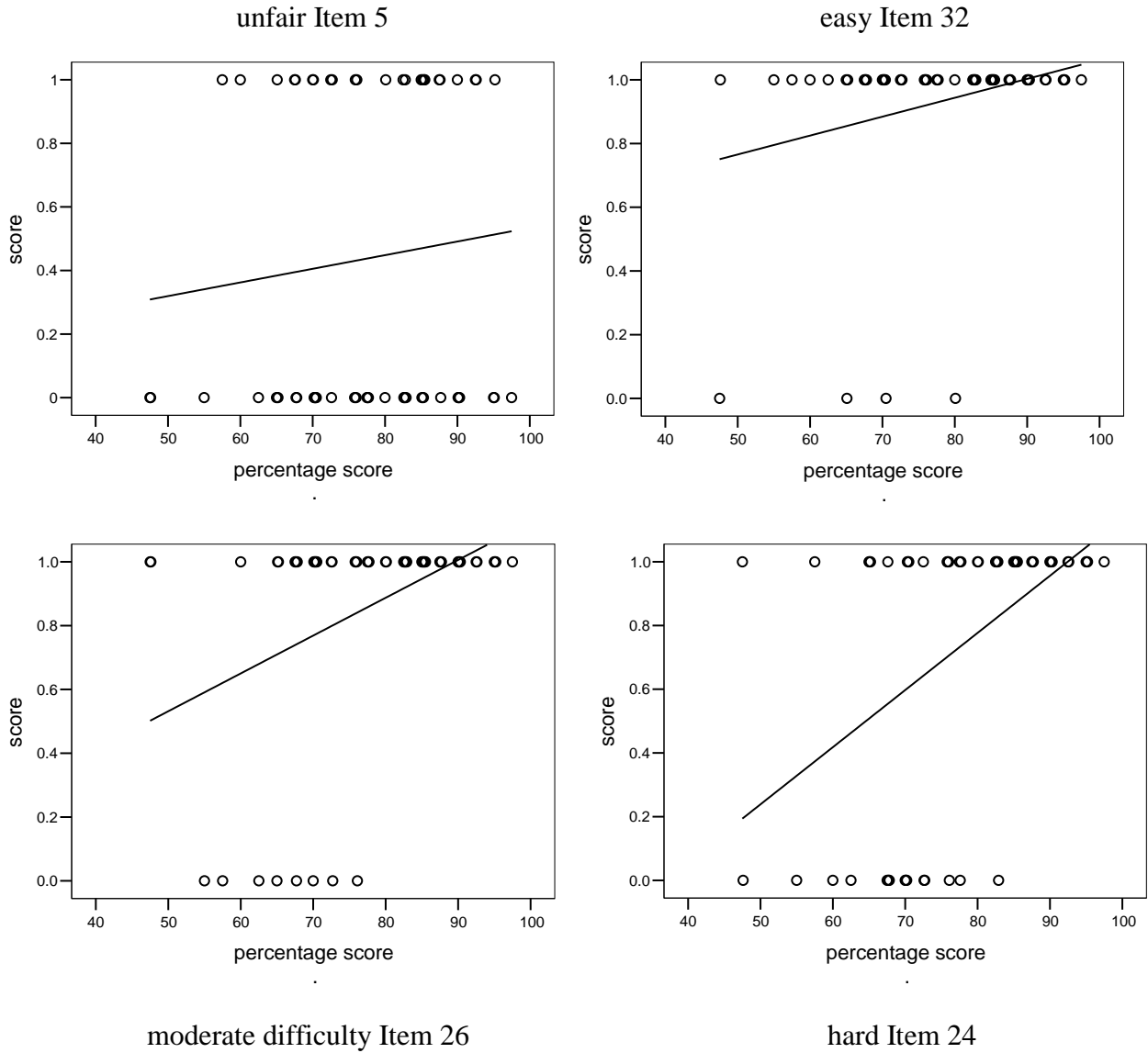

Fig.1. Item scores versus total percentage scores for four typical items and 55 examinees. The regression lines are characteristic of each item type.

We will use a normalized linear regression equation for an item

$$p(g) = b_0 + b_1 g, \qquad (1)$$

where $b_0$ and $b_1$ are the regression coefficients, $g$ denotes the normalized score of a student so that the percentage score equals $100g$ and $g=1$ means that the student got all the items right. $p(g)$ is the proportion of correct responses to the item for those examinees whose normalized score is $g$. The fit of the linear regression model to the empirical data was examined. Most of the data proved to be in close accord with the model.

Since Eq. (1) is defined by two parameters $b_0$ and $b_1$, we may interpret them as coordinates of a point $(b_0; b_1)$ in the $b_0$-$b_1$ plane. In Fig.2 all 40 test items are presented by such points in one graph.

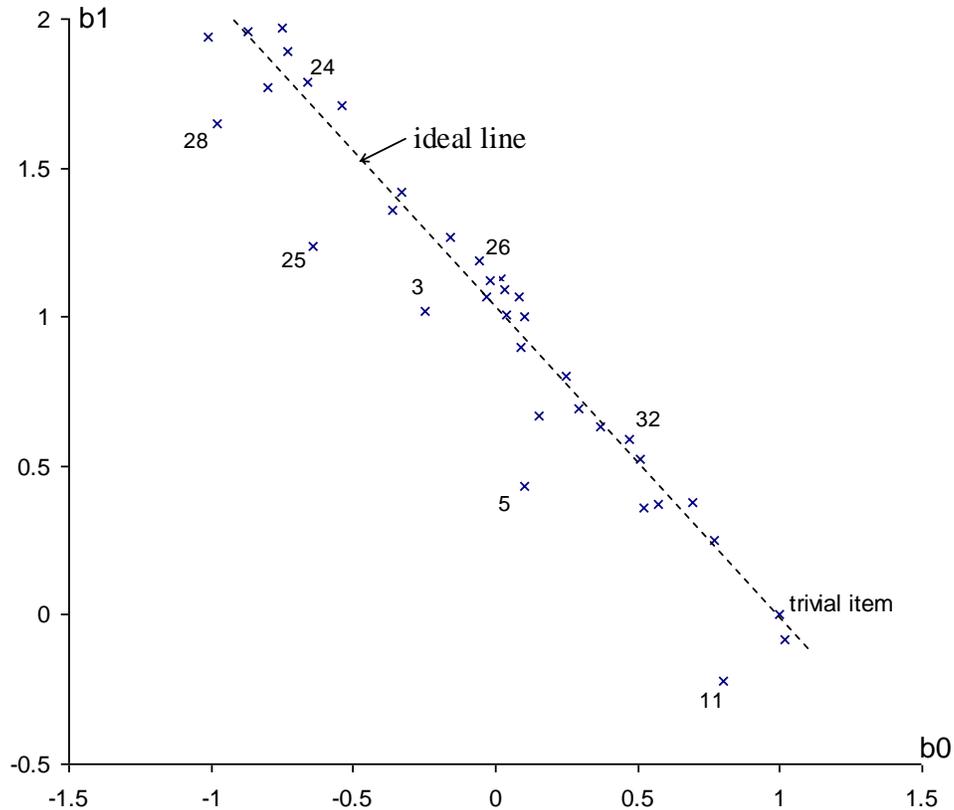

Fig.2. The graph shows that the points corresponding to items 5, 11, 25, and 28 are situated below and far from the ideal line.

## *2. Fair item definition*

The following three requirements to test items are natural:

1. The examinees have access to a detailed enumeration of all topics, special terms, and skills to be tested. We will call this enumeration a 'test specification'. A course with a recommended textbook may serve as a good test specification. Less detailed is a course without a textbook or a textbook without a course. Perhaps the least specified are licensing tests for an occupation.

2. The test item must be well worded and clear for all examinees.

3. For multi-choice tests the item must be correctly keyed.

We will call an item fair if all three conditions are satisfied for it.

In the light of the suggested item representation it is easy to distinguish between fair items and unfair ones. Provided the exam is made of perfectly fair items, the students with compete knowledge should respond correctly to all items. Now if we take any fair item, these students (whose normalized score $g=1$) will respond correctly to this specific item too, i.e. the proportion of correct responses $p(1)=1$. This means that for $g=1$ Eq. (1) turns into $1 = b_0 + b_1 \cdot 1$. In other words, in the ideal case all points $(b_0, b_1)$ of the fair items will be localized on a straight line whose equation is

$$b_0 + b_1 - 1 = 0. \qquad (2)$$

We will call this line 'the ideal line'. If all responses to an item are correct, we may label such an item as 'trivial'. The regression equation (1) for such an item transforms into $1=1+0\cdot g$, hence $(b_0, b_1) =(1, 0)$ is the point of all trivial items (see Fig.2). The ideal line goes through this point and the point $(0,1)$.

Unfair items are those for which Eq. (2) is violated. It may be an item that requires knowledge not supplied in the course or not included in the test specification. A badly worded item is also unfair as well as containing vague phrases. And of course, a wrongly keyed item is also unfair.

No matter the reason, not all successful students responded correctly to this item. It can be expressed mathematically by the regression equation (1) for the students with complete knowledge $p(1) = b_0 + b_1 \cdot 1$, where the proportion of correct responses $p(1)<1$ for an unfair item. The latter assertion means that the point corresponding to an unfair item in the graph does not lie on the ideal line but rather below it. Indeed, for such an item $b_0+b_1-1= b_0+b_1-p(1)+p(1)-1=p(1)-1 < 0$. Suppose no-one responded correctly to an item question, then $b_0=b_1=0$, $b_0+b_1-1= -1$ and the item is unfair.

Thus, we got a mathematical definition of an unfair item. One may see in Fig.2 that items 5, 11, 25, and 28 are problematic, however not all of the rest points lie on the ideal line. The reason for such a phenomenon is that the coefficients $b_0$ and $b_1$ are estimated statistically; moreover, in a non-ideal case unfair items are present. The occurrence of such items distorts students' scores, and also the regression coefficients of fair items. Hence the points of fair items on the graph drift from the ideal line. It is inevitable because (as it is shown in Appendix) the sum of the distances $d_i$ from the ideal line $\sum_i d_i = 0$. If a test includes unfair items (i.e. with $d_i<0$), then there must be items whose $d_i >0$.

Removing the unfair items 5, 11, 25, 28 from the current administration of the test brings the total scores closer to the true ones that in its turn will correct the regression coefficients. Before the unfair items deletion the sum of positive distances from the ideal line was 1.69, after the deletion it became 0.85, which is half of the previous value. Thus, the elimination of unfair items brings the fair item points closer to the ideal line.

## 3. Comparison with Item Response Theory (IRT)

Item Response Theory (e.g. Hambleton & Swaminathan, 1985) is similar to the suggested approach, since it presents nonlinear regression for the probability of responding correctly to an item. Alternatively the probability may be interpreted as the proportion of those who can answer the item correctly like in our approach. The regression function has the form

$$P(\theta) = c + \frac{1-c}{1+e^{-Da(\theta-b)}} \qquad (3)$$

where $\theta$ denotes a latent trait which underlies performance on a set of items, $a$ is a discrimination parameter for the item and $b$ is a difficulty parameter. The parameter $c$ is called the pseudo guessing parameter, $D=1.701$ is a constant.

The exponential wrapper of $P(\theta)$ is owed to the infinite range of $\theta$, however in practice it is final (from -3 to +3), thus in reality one might do without the complication. In our model the role of $\theta$ plays the total student score, thus the range is obviously limited, so the adjustment for infinite values is excessive. It is unclear whether IRT model applies to unfair items (fairness is not in the list of its assumptions), which makes an important difference between the two methods.

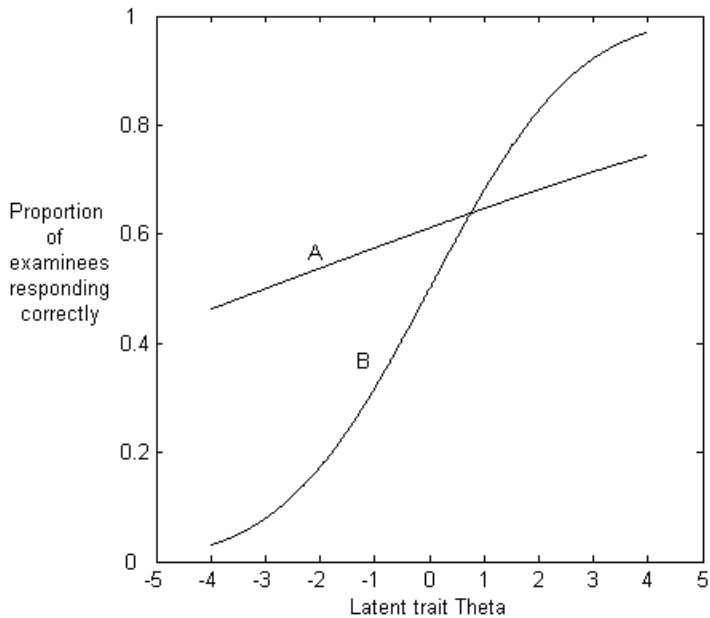

Figure 3. The Lord's paradox.

Indeed, consider an unfair item. If the function $P(\theta)$ describes it, then at some sufficiently great $\theta$ we must receive 100% correct responses for the geniuses, which is impossible in case of unfair item. No infinite talents will help to entirely cope with such items.

The Lord's paradox confirms the said. Figure 3 depicts two intercrossed curves P(θ) for two items of a test. For examinees whose θ is +3, the probability of scoring Item A is 0.7 while the probability of scoring Item B is 0.9. This means that Item A is the more difficult one for examinees of high ability. For examinees whose θ is -2, the probability of answering Item A correctly is 0.5 whereas the probability of giving the correct answer to Item B is 0.2. For them the latter item is harder. Thus, one cannot tell which of the two items is more difficult.

Our approach eliminates the Lord's paradox situation. As it was shown in the previous section, the regression line of any fair item passes through the point (1,1). If two straight lines intersect in two points, the lines coincide (see Figure 4). Therefore in the context of our approach the Lord's paradox is impossible for fair items. This situation appears in IRT, since it avoids mentioning unfair items, therefore it does not suggest a method for their determination. Burton (2006) also indicated that item response theory is inappropriate if the course covers many distinct topics which is typical for educational tests. For such tests the measure of internal consistency Cronbach's Alpha may be even negative (Cronbach and Shavelson, 2004).

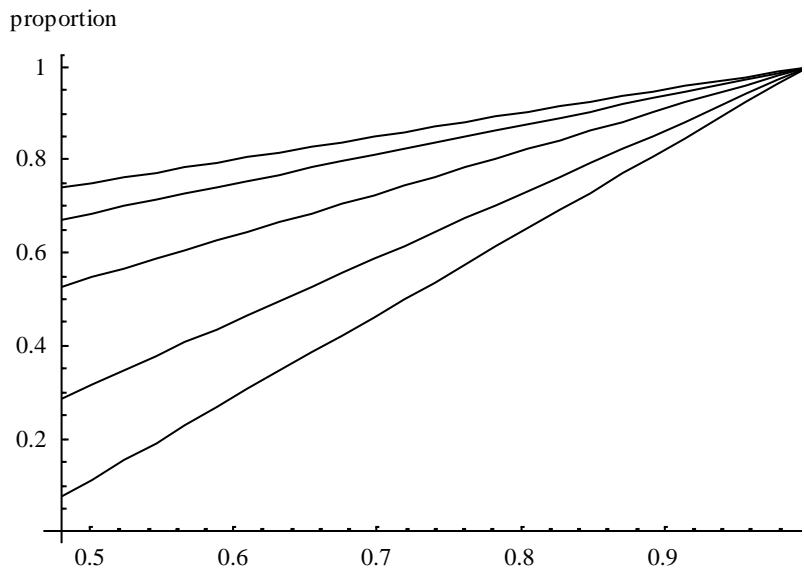

Figure 4. The Lord's paradox is impossible in the proposed method.

## 4. Procedure for unfair items identification

For the identification of unfair items it is necessary to calculate distances $d_i$ from each item point $(b_0, b_1)$ to the ideal line. According to the analytic geometry formula, the distance $d$ from a point $(x_0, y_0)$ to a straight line $Ax+By+C=0$ is $d = \dfrac{Ax_0 + By_0 + C}{\sqrt{A^2 + B^2}}$. For our case A=B=1, C= −1, hence $d = \dfrac{b_0 + b_1 - 1}{\sqrt{2}}$. Since the unfair item points are situated below the ideal line, their distances $d_i$ are negative.

The whole process of calculations was automated with the help of the computer program Consensus5 (Bakman, 2007). The consensus method identifies erroneous data, defective measuring instruments, and contaminated distributions. The latter option fits our case. Indeed, if one considers the distance $d$ a random quantity, then the unfair item distances $d_i$ will gather near one end of its distribution. In contrast, fair item distances are close to zero, hence they form a consensus which allows identifying unfair items lying outside the consensus.

The routine Consensus5 calculates the coefficients $b_0^{(i)}$, $b_1^{(i)}$ for each item $i$, as well as the distance $d_i$ from the ideal line, and also the limit for fair item distance from the ideal line $d_f$ ($d_i > d_f$ means that $i$-th item is unfair). Based on this value Consensus5 produces the list of unfair items. After this the routine removes the latter and repeats the process. The cycle stops if no items were removed. For the discussed test, the routine Consensus5 identified the items 5, 11, 25, and 28 as unfair, the corresponding distances from the ideal line being -0.33, -0.29, -0.29, -0.23.

## 5. Comparison with other methods of unfair item identification

Many item analysis manuals recommend discarding items with negative correlations between the item scores and the total score (e.g. Oosterhof, 2001). We will call this approach 'traditional'. Figure 5 repeats Fig.2 with the difference, that it illustrates similarity and distinction between the traditional and the proposed methods for unfair item identification.

First consider the traditional method. The correlation coefficient $r$ and $b_1$ are interconnected $b_1 = r \cdot s_p / s_g$, therefore if $r < 0$ then $b_1$ is also negative. Thus, the region of unfair items (from the traditional method viewpoint) lies below the straight line $b_1=0$. According to the proposed method, unfair item points lie below the ideal line. Moreover, their distances from the ideal line are greater than the cutoff value $d_f$. For our example $d_f = 0.2$, this boundary is marked by the line *AB* in Fig.5. Since p(1)=$b_0+b_1 \cdot 1 \geq 0$, unfair item points cannot lie below the line $b_0+b_1=0$, this bottom boundary is marked in Fig.5 by the line *CD*. Thus, according to our approach, unfair item points are located in the strip between the lines *AB* and *CD*.

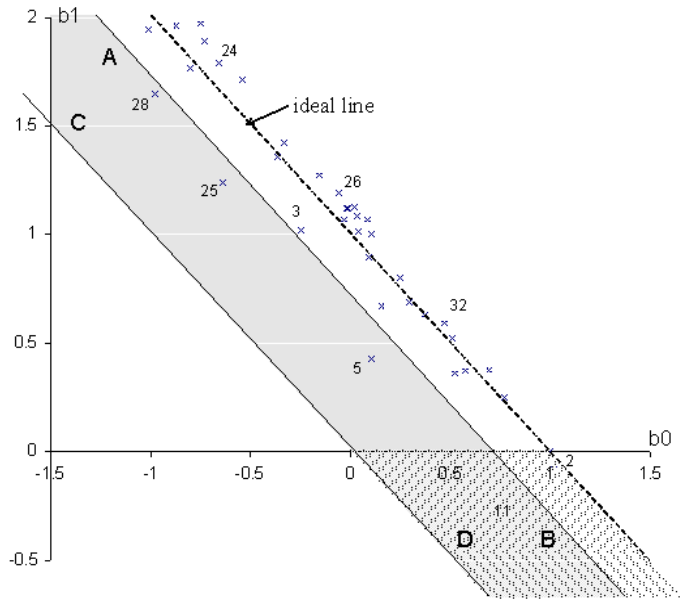

Figure 5. According to the traditional method (based on *r*<0 condition), unfair item points lie in the sandy region, whereas the proposed method identifies items as unfair if their points lie in the shaded strip between the lines *AB* and *CD*. Thus, the sandy-shaded region is common for both methods.

As a result three regions are formed: the shaded region where item points, unfair only for our method, lie; the common sandy-shaded region, in which lie item points unfair for both methods. The sandy (non-shaded) region contains item points unfair only for the traditional method.

Lecturer B's group has representatives in all three regions. In the shaded region there are three items 5, 25 and 28 which have been determined as unfair by the proposed method. This conclusion was supported by experts' judgments. These items remain unrevealed by the traditional method, because their item-total correlations are positive. For our exam we have

| item | correlation |
|------|-------------|
| 5    | 0.10        |
| 25   | 0.32        |
| 28   | 0.43        |

Consider for example Item 25. Its topic was insufficiently taught to the students which made it unfair ($d_i = -0.29$), however its item-total correlation is high ($r=0.32$), hence the item is fair according to the traditional method. Thus, the latter cannot reveal shortcomings of teaching.

Item 11 lies in the sandy-shaded region, hence it is recognized as unfair by both methods.

Item 2 belongs to the sandy (non-shaded) region, which is controversial: the item is almost trivial (53 of 55 students got it right) and a trivial item cannot be unfair. The same conclusion follows from our criterion: Item 2 point lies near the ideal line, hence the item is fair. In contrast, the item-total correlation $r = -0.05$, consequently, according to the traditional method the item is unfair. Another example of faulty item identification in the sandy region occurs if a test measures many different content areas and cognitive skills. Then a fair item, assessing special ability, may have a low item-total correlation, so it may be identified as unfair by the traditional method.

Actually *r* is one form of item discrimination index (Burton, 2001). Mehrens and Lehmann (1973, p.333-4) provide the following set of cautions in using item analysis: 'The discrimination index is not always a measure of item quality (emphasis is ours). There is a variety of reasons an item may have low discriminating power:

(a) extremely easy items will have low ability to discriminate but such items are often needed to adequately sample course content and objectives;

(b) an item may show low discrimination if the test measures many different content areas and cognitive skills. For example, if the majority of the test measures "knowledge of facts", then an item assessing "ability to apply principles" may have a low correlation with total test score, yet both types of items are needed to measure attainment of course objectives.'

The performed analysis shows that *r* is clearly insufficient for the identification of unfair items.

*Differential Item Functioning* (*DIF*) is another method for unfair item detection.

DIF verifies whether a test item is biased against particular gender, ethnic, social, or economic groups (see, for example, Holland, 1988). DIF tries to assess item functioning by comparison of the item scores for two or more groups. For our method it is not necessary to compare groups to identify unfair items – one group is enough. The distinction becomes more obvious, if we consider an item unfair for all groups, for example, a vaguely worded item. Our method will identify such unfair item with and without group separation, while Differential Item Functioning cannot reveal any bias, because comparison of groups will not show any difference between them.

Another problematic situation in which DIF cannot detect item bias occurs if the test item measures two skills, one of which is biased, for example, against male, but the second one is biased against female. Then DIF is also helpless.

## 6. How teaching influences problematic item scores

Since exam scores are influenced by instructional procedures employed, it is only natural to expect that teaching might affect at least certain categories of problematic items. In order to verify the hypothesis we used responses of two additional groups of students that took the same course and passed the same exam. Their lecturers were D and L.

There is distinction in the item unfairness between the groups, e.g. in D's group Item 3 is unfair instead of Item 11 in B's group, nevertheless there is agreement about the unfairness of items 5, 25, and 28.

Table 2 presents distances $d_i$ from the ideal line for five items which were determined as unfair at least in one of the three groups. The data are given separately for each group of teaching and for all three groups together.

| Lecturer | Items that were determined as unfair at least in one of the groups | | | | |
|---|---|---|---|---|---|
| | 3 | 5 | 11 | 25 | 28 |
| B | -0.12 | **-0.33** | **-0.29** | **-0.29** | **-0.23** |
| D | **-0.38** | **-0.30** | 0.04 | **-0.39** | **-0.18** |
| L | -0.01 | **-0.21** | 0.05 | **-0.56** | -0.12 |
| all three groups | **-0.16** | **-0.27** | -0.05 | **-0.44** | **-0.17** |

Table 2. Distances $d_i$ from the ideal line for various groups of teaching are presented for five items which were determined as unfair at least in one of the three groups. Distances for items that were not determined as unfair in the corresponding group are printed in a small font.

The unfair items are characterized as follows. Item 3 incorporated a non-accurate mathematical formulation; Item 5 included a vague phrase. One term used in Item 11 was unfamiliar for some students; the topic of Item 25 was insufficiently taught in class; Item 28 had overburdened formulation. Though it might seem that items 3 and 5 are unfair irrespective of the teacher, in essence the teacher could accustom his/her students to utilization and to perception of the obscure phrase of Item 5 and even to the non-accurate mathematical formulation of Item 3.

## *7. Generalization of the method for any exam*

Such generalization may be achieved by normalizing each item score so that the maximum score amounted to 1 for any item. The normalized score is obtained by dividing the usual score by the maximum possible score for the item. For instance, if the maximum score is 5, then the normalized scores will be 1/5 for the score 1, 2/5 for the score 2 and so on. As a result all items become equally weighted, however after the identification and elimination of unfair items, the educator may regain the item weights in computing the final scores.

After the normalization we come to the situation similar to the described in Section 1, except for additional options of fractional item scores. Thus, the equation of the ideal line and all the conclusions remain as before.

It should be noted that no assumptions were made about the normality of the score distributions as it is usually accepted, neither about their error independence, so that the proposed method has the highest domain of applicability.

## *8. Summary*

The proposed method transforms students' scores into the unique measure, namely the distance from the item point to the ideal line. If all items of a test are fair, the distances should be equal to zero, however for real exams the fair item points are localized near the ideal line. Points representing unfair items are situated far from the ideal line; they are outside of the consensus, therefore the consensus method identifies them.

The suggested method has advantages over other methods for unfair item identification. It was shown that item-total correlations are insufficient for identification of unfair items because there exist unfair items with positive correlations and fair items with negative ones. Differential Item Functioning (DIF) does not reveal unfair items, if they are evenly unfair in subgroups or if the test item measures two skills biased in opposite directions. The proposed method manages successfully all these cases.

Though the proposed mathematical definition of unfair items may be applied only after an exam, nevertheless their removal and consequent recalculation of students' scores will make the exam results trustier. Thereto, the identification of unfair items gives the educator instructional feedback for developing and refining items. Without such a feedback the educator might replicate the same errors again and again.

## *Acknowledgments*

The author would like to thank Richard Burton and Michael Zieky for reading the manuscript and providing most valuable comments.

Appendix. Additional analysis

Since values of $p(g)$, $b_0$, $b_1$ in (1) vary over items on a test, for the additional analysis we will write them as $p_i(g)$, $b_0^{(i)}$, $b_1^{(i)}$, where $i$ denotes $i$-th item on a test. The regression equation for the $i$-th item appears in the following form

$$p_i(g) = b_0^{(i)} + b_1^{(i)} \cdot g \qquad (4)$$

where $g = \sum_{i=1}^{N} x_{ki} / N$, $N$ is the number of items. Let us sum up (4) for $i$ from 1 to $N$. We obtain

$$\sum_i p_i(g) = \sum_i b_0^{(i)} + g \cdot \sum_i b_1^{(i)} \qquad (5)$$

According to the definition $p_i(g) = \sum_{k \in K_g} x_{ki} / K_g$, where $K_g$ denotes the set of indices $k$ of those students whose grade is $g$. It also denotes here their quantity without causing ambiguity. Summing up $p_i(g)$ for $i$ from 1 to N and exchanging the order of summation yields

$$\sum_i p_i(g) = \frac{1}{K_g} \sum_{k \in K_g} \sum_i x_{ki} = \frac{1}{K_g} \sum_{k \in K_g} Ng = Ng$$

Substituting the latter in (5) yields $Ng = \sum_i b_0^{(i)} + g \sum_i b_1^{(i)}$. Therefore

$$g = \bar{b}_0 + g \cdot \bar{b}_1, \tag{6}$$

where $\bar{b}_0$ and $\bar{b}_1$ are the averaged regression coefficients over all items in the test.

Now (6) may be rewritten in the form $\bar{b}_0 + g \cdot (\bar{b}_1 - 1) = 0$. Taking into account that this equality holds for all values of $g$, yields $\bar{b}_1 = 1$. But then $\bar{b}_0 = 0$ which means that the midpoint of the graph for all items of any test is the point (0,1). Since (6) is an exact identity, $\bar{b}_1 = 1$ as well as $\bar{b}_0 = 0$ hold even for tests containing only unfair items. Items whose points are neighbors of (0,1) may serve as representatives of the average item difficulty across the test.

Using the received result, it can be shown that for any test $\sum_i d_i = 0$. Indeed,

$\frac{1}{N} \sum_i d_i = \frac{1}{N} \sum_i \frac{b_0^{(i)} + b_1^{(i)} - 1}{\sqrt{2}} = \frac{1}{\sqrt{2}} (\bar{b}_0 + \bar{b}_1 - 1) = 0$. This result was utilized in Section 2.

We saw in our illustrative example that $b_1$-values run from 0 (for trivial items) to 1 for the moderate difficulty items and then to 2 for hard ones. Thus, one may use the regression coefficient $b_1$ as a measure of item difficulty. Since $b_1$ is the linear regression line slope, it may also serve as a discriminator between those who scored high on the total test and those who scored low.